\documentclass[sigconf,nonacm]{acmart}

\setcopyright{none}                
\acmDOI{}
\acmISBN{}
\acmConference{}{}{}
\acmBooktitle{}
\acmYear{}
\acmMonth{}
\acmPrice{}

\usepackage{algorithmicx}
\usepackage{algorithm}
\usepackage{enumitem}
\usepackage{booktabs}
\usepackage{multirow}

\title{JEDA: Query-Free Clinical Order Search from Ambient Dialogues}

\author{Praphul Singh}
\affiliation{%
  \institution{Oracle Health \& AI}
  \city{}
  \country{}}
\email{praphul.singh@oracle.com}

\author{Corey Barrett}
\affiliation{%
  \institution{Oracle Health \& AI}
  \city{}
  \country{}}
\email{corey.barrett@oracle.com}

\author{Sumana Srivasta}
\affiliation{%
  \institution{Oracle Health \& AI}
  \city{}
  \country{}}
\email{sumana.srivasta@oracle.com}

\author{Amitabh Saikia}
\affiliation{%
  \institution{Oracle Health \& AI}
  \city{}
  \country{}}
\email{amitabh.saikia@oracle.com}

\author{Irfan Bulu}
\affiliation{%
  \institution{Oracle Health \& AI}
  \city{}
  \country{}}
\email{irfan.bulu@oracle.com}

\author{Sri Gadde}
\affiliation{%
  \institution{Oracle Health \& AI}
  \city{}
  \country{}}
\email{sri.gadde@oracle.com}

\author{Krishnaram Kenthapadi}
\affiliation{%
  \institution{Oracle Health \& AI}
  \city{}
  \country{}}
\email{krishnaram.kenthapadi@oracle.com}

\begin{document}
\begin{abstract}
Clinical conversations interleave explicit directives ("order a chest X ray") with implicit reasoning in ambient dialogue ("the cough worsened overnight, we should check for pneumonia"). Many retrieval systems insert an intermediate LLM to rewrite such inputs into structured queries, which adds latency, instability, and opacity that limit real time clinical ordering. We present \textbf{JEDA} (Joint Embedding for Direct and Ambient clinical orders), a framing in which a domain initialized bi-encoder retrieves canonical orders directly from conversational text without query rewriting and, in a query free mode, encodes a short rolling window of ambient dialogue to trigger retrieval without explicit query formulation. Initialized from PubMedBERT and fine tuned with a duplicate safe contrastive objective, the model aligns heterogeneous expressions of intent to shared order concepts. Training records are produced via constrained LLM guidance that ties each signed order to complementary formulations, namely command only, context only (ambient), command+context, and context+reasoning, capturing how clinicians naturally express intent. This variant mixed supervision yields a unified embedding space with clearer inter order separation, tighter query to order coupling, and stronger generalization across phrasing styles. In practice the query free framing is noise resilient, reducing sensitivity to conversational disfluencies and ASR errors by conditioning on a short rolling window rather than a single utterance while avoiding brittle query formulation. When deployed, the approach delivers large gains in retrieval accuracy and latency over rewriting based pipelines and substantially outperforms both its base encoder (PubMedBERT) and recent open embedders including Linq Embed Mistral, SFR Embedding, GTE Qwen, BGE large, and Embedding Gemma across all evaluation conditions. The result is a fast, interpretable, and LLM free retrieval layer that links ambient conversational context to actionable clinical orders and operates query free in real time.
\end{abstract}

\maketitle



\section{Introduction}

Clinical conversations blend two distinct linguistic registers. Providers issue direct commands that map cleanly to orderables (for example, "order chest X ray"), and they carry ambient dialogue about symptoms, history, and rationale that only implicitly points to an order (for example, "the cough worsened overnight, we should check for pneumonia"). State of the art sentence encoders excel at semantic matching and retrieval \cite{reimers2019sentence, gao2021simcse, ni2021sentence}, yet most healthcare oriented encoders are tuned either to narrative notes or to structured strings, not to both simultaneously \cite{alsentzer2019publicly, huang2019clinicalbert, neuml2020pubmedbert, lee2020biobert, shin2020biomegatron}. In practice, many systems insert a large language model to rewrite conversation into a canonical query before retrieval \cite{lewis2020rag, nogueira2019doc2query, ma2023queryrewriting, lin2023llmretrieval}. This extra hop introduces latency, increases variability under underspecified or ambiguous input, and complicates provenance in real time clinical workflows. Real time order suggestions are particularly sensitive to interaction latency and variability given the documentation burden clinicians already face in EHR use \cite{sinsky2016allocation,shanafelt2016clerical}. In addition, spontaneous clinical speech contains disfluencies and is susceptible to ASR errors, which further destabilize query rewriting pipelines \cite{shriberg2001disfluencies,gulati2020conformer,radford2023whisper}.

Bi-encoder retrieval offers a compelling alternative: it separates encoding from search, scales to large indices, and delivers sub second latency \cite{karpukhin2020dense, xiong2021approximate, izacard2021contriever, thakur2021beir}. However, standard contrastive training typically assumes a single query form. Clinical encounters, by contrast, present heterogeneous expressions of the same underlying intent, ranging from terse commands to long contextual descriptions with implicit rationale. Moreover, in batch negatives can become false negatives when multiple queries in a minibatch point to the same order, which degrades the learned geometry if unaddressed.

This work targets the problem of retrieving canonical orders directly from conversational text without an intermediate rewriter and adds a query free operating mode that uses a short rolling window of ambient dialogue as the retrieval input. The approach is a Bi-encoder that aligns conversational fragments with canonical orders using a duplicate safe contrastive objective. We initialize from PubMedBERT to inherit biomedical vocabulary and domain priors \cite{neuml2020pubmedbert, gu2021pubmedbert}, then fine tune with a Multiple Negatives Ranking loss that masks duplicate positives within a minibatch so that queries sharing the same target order are not treated as negatives \cite{reimers2019sentence, gao2021simcse}. Training records are constructed with constrained LLM guidance that ties together imperative commands, verbatim context spans, and concise reasoning for each signed order, yielding variant rich supervision that reflects how order intent appears in real conversations \cite{yang2022generative, jin2021deep}. This formulation encourages a shared embedding space in which heterogeneous expressions converge on the same canonical concept while preserving fine grained distinctions among closely related orders.

The design complements retrieval augmented generation pipelines by strengthening the retriever at the conversation interface rather than relying on query reformulation \cite{lewis2020rag, ma2023queryrewriting, lin2023llmretrieval}. It builds on advances in dense retrieval and contrastive sentence representation learning \cite{karpukhin2020dense, xiong2021approximate, izacard2021contriever, reimers2019sentence, gao2021simcse} and follows best practices in domain adaptation for clinical NLP \cite{alsentzer2019publicly, huang2019clinicalbert, gu2021pubmedbert}. Unlike biomedical entity linking methods centered on UMLS names \cite{sung2020biosyn, liu2021sapbert, bodenreider2004umls}, the focus here is conversation to order alignment and robustness across direct and ambient forms.

We evaluate the approach on a temporally held out set of encounter derived queries using a unified order corpus. Retrieval metrics and embedding structure diagnostics indicate that the learned space exhibits larger inter order margins, tighter query to order coupling, and improved ranking calibration compared to a strong PubMedBERT baseline. Qualitative UMAP visualizations corroborate these trends, showing compact neighborhoods around the correct order embeddings and clearer boundaries between distinct orders. The same encoder operates in a query free mode when the explicit query is replaced by a short ambient dialogue window, yielding the same robustness to underspecified inputs observed for ambient variants. Together, these results suggest that a simple duplicate safe contrastive recipe over variant aware supervision can reduce reliance on intermediate rewriting while better aligning natural dialogue with structured orderables in clinical settings. The same encoder serves both explicit query and query free ambient inputs without architectural changes, simplifying deployment.

\section{Related Work}

\textbf{Conversational Retrieval and RAG.}
Retrieval augmented generation integrates dense retrieval with generation models, grounding outputs in external corpora \cite{lewis2020rag, izacard2021contriever}. In clinical contexts, conversational RAG systems have been explored for summarization and knowledge grounding \cite{yang2022generative, jin2021deep}. Recent work examines real time speech summarization for medical conversations \cite{leduc2024realtimesum} and knowledge graph aided medical RAG \cite{medrag2025}, expanding the design space for clinically oriented retrieval pipelines. Beyond note centric corpora, large scale medical dialogue datasets highlight the variability and spontaneity of patient and provider conversations \cite{zeng2020meddialog}. However, most designs rely on LLM based query reformulation to translate raw dialogue into structured queries \cite{lin2023llmretrieval, nogueira2019doc2query}. This adds semantic instability when conversations are vague or underspecified and also adds latency since additional LLM calls are required. Such architectures are ill suited for real time clinical decision support where order proposals must be generated synchronously during encounters. By contrast, a query free framing uses a short ambient dialogue window as the retrieval input, avoiding explicit query formulation while preserving provenance. A query free windowed approach preserves provenance by operating directly on conversational evidence rather than synthetic rewrites.

\textbf{Embedding Models for Retrieval.}
Dense retrievers such as DPR \cite{karpukhin2020dense}, ANCE \cite{xiong2021approximate}, and Contriever \cite{izacard2021contriever} established Bi-encoder architectures as state of the art in open domain QA and passage search. Sentence level embeddings such as Sentence BERT \cite{reimers2019sentence}, SimCSE \cite{gao2021simcse}, and Sentence T5 \cite{ni2021sentence} extended contrastive learning for semantic similarity and clustering. Metric learning frameworks incorporating hard negatives further improved discriminative power \cite{schroff2015facenet, musgrave2020metric}, with additional perspectives from multi similarity losses \cite{wang2019multisim}, contrastive predictive coding \cite{vandenoord2018cpc}, and neighborhood components analysis \cite{goldberger2005nca}. While effective in general domains, these approaches are not optimized for conversational medical dialogue.

\textbf{IR Benchmarks and Infrastructure.}
The BEIR benchmark provides a heterogeneous suite for zero shot and transfer evaluation of retrieval models across diverse tasks and domains \cite{thakur2021beir}. From a systems perspective, FAISS enables billion scale vector indexing and approximate search on CPUs and GPUs \cite{johnson2019billion}. The proposed Bi-encoder retriever is compatible with such infrastructure and serves from standard ANN indexes while specializing the training objective for dialogue to order alignment. Operationally, low latency serving often combines FAISS with graph based indexes such as HNSW or uses benchmarking guided configurations to meet tight throughput constraints \cite{johnson2019billion,malkov2018hnsw,aumuller2017annbenchmarks}.

\textbf{Clinical and Biomedical Embeddings.}
Domain adapted models such as ClinicalBERT \cite{huang2019clinicalbert}, BioBERT \cite{lee2020biobert}, PubMedBERT \cite{neuml2020pubmedbert}, and Med BERT \cite{rasmy2021medbert} improved performance on clinical classification, similarity, and inference. Studies have emphasized the importance of domain specific corpora \cite{alsentzer2019publicly, beam2020clinical}. Biomedical retrieval benchmarks such as MedSTS \cite{medsts2018} and MedNLI \cite{romanov2018mednli} further standardize evaluation. However, these models treat narratives or structured fields in isolation, without unifying dialogue and order intent.

\textbf{Biomedical Entity Linking and Self Alignment.}
A parallel line of work aligns surface forms to canonical biomedical concepts using metric learning over UMLS. BioSyn \cite{sung2020biosyn} introduces synonym marginalization and iterative candidate retrieval; SapBERT \cite{liu2021sapbert} self aligns synonyms at scale with UMLS supervision \cite{bodenreider2004umls}, leveraging MS loss and online hard example mining \cite{wang2019multisim}. Datasets and toolkits such as MedMentions \cite{mohan2019medmentions}, COMETA \cite{basaldella2020cometa}, and SciSpacy \cite{neumann2019scispacy} catalyzed progress, with recent surveys standardizing evaluation \cite{kartchner2023bioel}. Our work departs from entity linking by repurposing PubMedBERT’s biomedical initialization to bridge conversational evidence and order concepts, specializing the embedding space with a duplicate safe contrastive objective.

\textbf{Order Retrieval and Clinical Applications.}
Retrieval oriented healthcare applications include ontology alignment \cite{zhang2020concept}, medical IR benchmarks \cite{thakur2021beir}, and trial matching systems such as Trial2Vec \cite{zhang2022trial2vec}. Embedding based methods have been applied to structured EHRs \cite{rasmy2021medbert, beam2020clinical}, but the task of retrieving canonical orders directly from conversational evidence and operating in a query free ambient mode has not been systematically addressed. The present work proposes a unified embedding framework bridging ambient dialogue and directive order commands, enabling fast and accurate order retrieval without intermediate LLM reformulation.

In contrast to prior art, this work directly addresses the duality of clinical language by creating joint embeddings for indirect conversational evidence and direct order intent while supporting a query free trigger from ambient dialogue. This fills a gap between biomedical embedding research and practical order retrieval in conversational clinical settings.

\section{Methods}

\subsection{Problem Setup}
Each clinical encounter $e$ consists of two sources of information:
(i) a transcript segmented into chunks
$U_e = \{u_{e,1}, \dots, u_{e,n_e}\}$, and
(ii) the set of signed orders
$S_e = \{s_{e,1}, \dots, s_{e,m_e}\}$.
Unlike approaches that rely on explicit timestamps or structured EHR links, no temporal alignment is available between conversation and orders. The learning task is to infer which portions of the dialogue motivated which orders, and to construct embeddings that jointly align conversational evidence with canonical order concepts.

\subsection{LLM-Guided Record Construction}
\begin{figure}[t]
  \centering
  \includegraphics[width=0.85\linewidth]{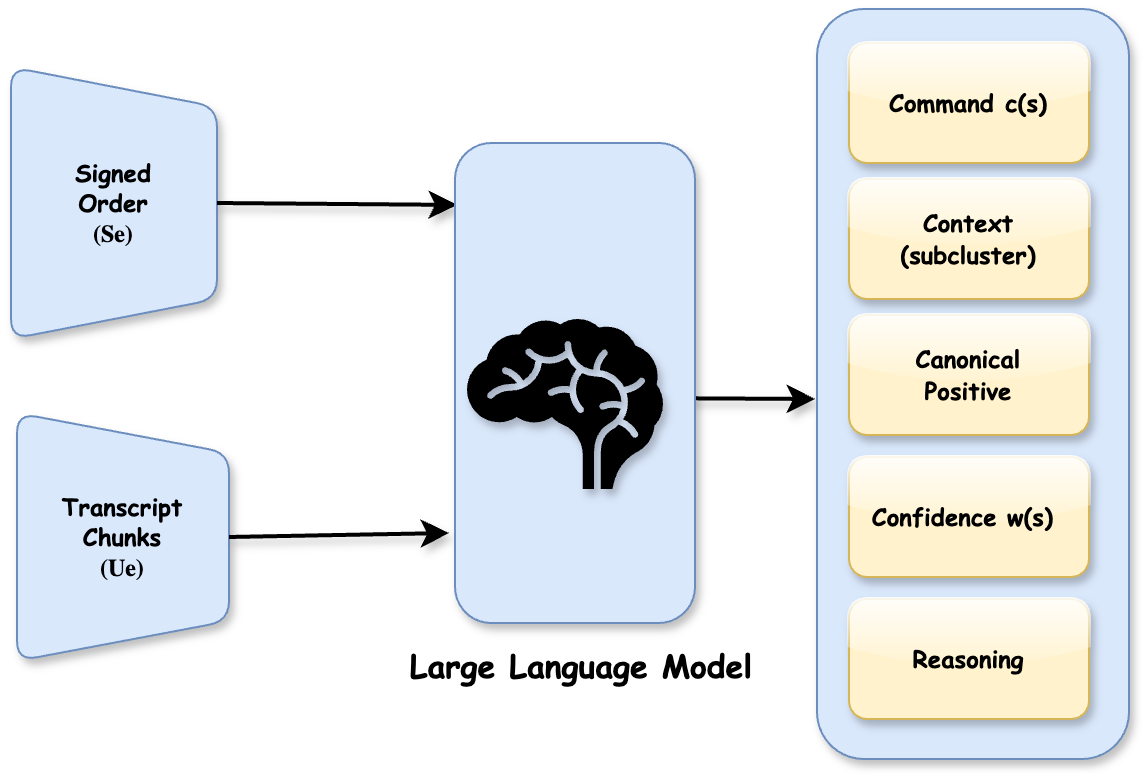}
  \caption{Training record construction. For each signed order, the pipeline selects supporting transcript subclusters, synthesizes a command, preserves verbatim context, generates a concise reasoning, and maps to a canonical concept.}
  \label{fig:record-construction}
\end{figure}

As shown in Figure~\ref{fig:record-construction}, a constrained large language model bridges transcripts and orders by producing structured training records. For each signed order $s \in S_e$, the model receives the full transcript $U_e$ and performs five steps.

\emph{First}, it selects a supporting subcluster $C(s) \subseteq U_e$ of utterance chunks that justify placing the order.
\emph{Second}, it synthesizes a directive $c(s)$ from these chunks, yielding an imperative command such as ``Order a chest X ray".
\emph{Third}, it preserves verbatim evidence by concatenating the supporting spans:
\[
\text{context}(s) = \texttt{" "}.join\{u : u \in C(s)\}.
\]
\emph{Fourth}, it generates a short \emph{reasoning} string $r(s)$ that explains, in natural language, why the order is appropriate given the context.
\emph{Finally}, the order string $s$ is mapped to a canonical ontology concept $o^+(s)$. In practice, canonical order concepts are typically grounded in standard clinical terminologies such as SNOMED CT, LOINC, and RxNorm \cite{donnelly2006snomed,vreeman2016loinc,nelson2011rxnorm}.

Each record includes a confidence score $w(s) \in [0,1]$ emitted by the LLM prompt that judges the strength of support. This confidence is not used in training but can guide data filtering.

\subsection{Query Variants}
To expose the model to both abstract and concrete signals, we derive four query formulations from each record:
\[
q_{\text{cc}} = \texttt{COMMAND: } c(s) \;\; \texttt{CONTEXT: } \text{context}(s)
\]
\[
q_{\text{c}} = \texttt{COMMAND: } c(s)
\]
\[
q_{\text{ctx}} = \texttt{CONTEXT: } \text{context}(s)
\]
\[
q_{\text{ctx+r}} = \texttt{CONTEXT: } \text{context}(s) \;\; \texttt{REASONING: } r(s)
\]
For example, if the transcript contains "I have burning with urination," the LLM may generate a command "Order a urinalysis," a context string quoting the symptom, and a brief reasoning such as "Urinalysis is indicated to evaluate dysuria and rule out infection." The combined query $q_{\text{cc}}$ fuses intent and evidence, $q_{\text{c}}$ and $q_{\text{ctx}}$ isolate intent and evidence respectively, and $q_{\text{ctx+r}}$ augments ambient evidence with explicit rationale.

We formulate multiple query variants to mirror how clinical intent appears in practice and to inject complementary learning signals. The \emph{command only} form isolates explicit intent and teaches the encoder to distinguish the target order from near neighbors even without local evidence. The \emph{context only} form captures purely ambient dialogue and encourages retrieval directly from transcript spans without an explicit directive, enabling query free triggering. The combined \emph{command+context} form ties intent to local evidence and reduces ambiguity, training the model to privilege grounded matches over surface similar but unsupported candidates. The \emph{context+reasoning} form makes the link between evidence and order explicit, pushing the representation to encode causal sufficiency rather than keyword co occurrence. Training across these variants regularizes the space against shortcut learning, improves robustness to phrasing differences across speakers and settings, and preserves cross variant invariance by aligning all variants to the same canonical order under the duplicate safe contrastive objective.

\subsection{Base Encoder and Initialization}
We use a single Transformer encoder initialized from \emph{PubMedBERT} \cite{neuml2020pubmedbert}. The encoder is shared across the two towers (tied weight Bi-encoder, $f_Q \equiv f_D \equiv f$) and produces $\ell_2$ normalized embeddings for both queries and documents. Biomedical grounding provides a strong prior for terminology, synonymy, and phrase variation, helping bridge conversational phrasing and canonical order strings.

\begin{figure}[t]
    \centering
    \includegraphics[width=\linewidth]{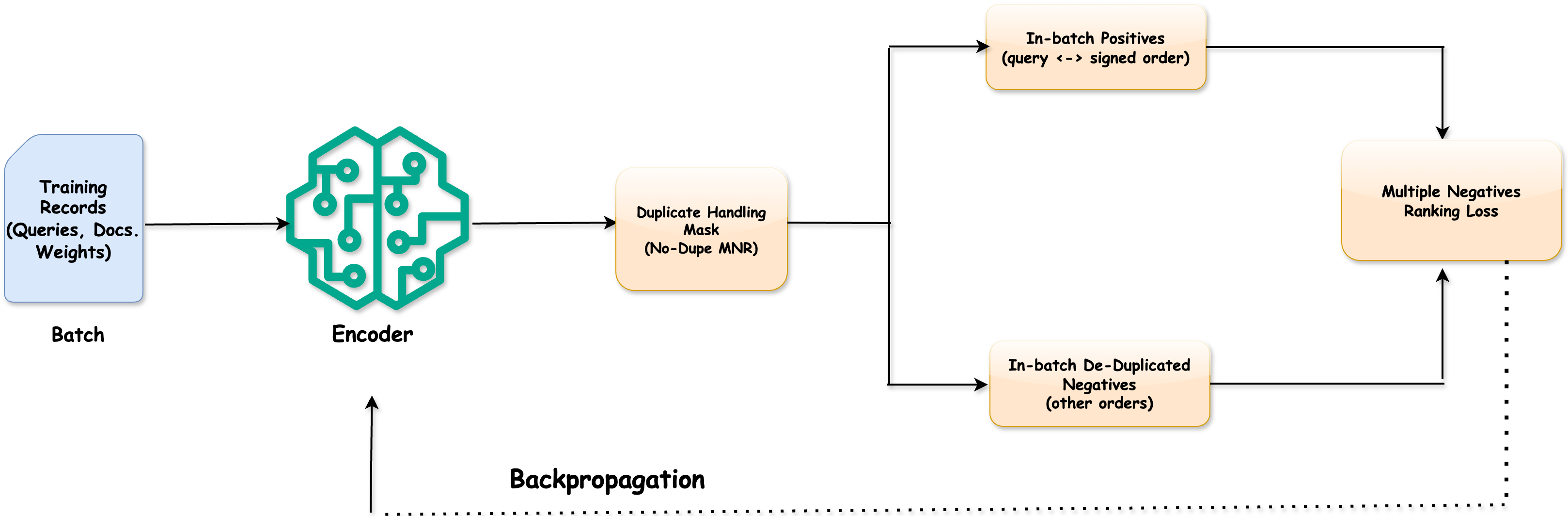}
    \caption{Contrastive fine tuning with duplicate safe Multiple Negatives Ranking. Queries and documents are encoded with a tied Bi-encoder, duplicate positives are masked, and the MNR loss is applied.}
    \label{fig:phase1}
\end{figure}

\subsection{Contrastive Training Objective}
As illustrated in Figure~\ref{fig:phase1}, the encoder is fine tuned to align conversational queries with signed orders using a duplicate safe Multiple Negatives Ranking loss. We train a tied Bi-encoder with $f_Q \equiv f_D \equiv f$, mapping inputs to $\ell_2$ normalized vectors in $\mathbb{R}^d$ so dot products correspond to cosine similarities. For a minibatch $B=\{(q_i, d_i^{+})\}_{i=1}^{|B|}$, the per example loss is
\begin{align}
\ell_i &=
- \log \frac{
   \exp\!\big(\tfrac{1}{\tau}\, f(q_i)^\top f(d_i^+)\big)
}{
   \sum_{j=1}^{|B|}
   M_{ij}\, \exp\!\big(\tfrac{1}{\tau}\, f(q_i)^\top f(d_j^+)\big)
},
\end{align}
where $\tau$ is a temperature and $M_{ij}\in\{0,1\}$ masks duplicate positives so that queries targeting the same order are not treated as negatives. This masking directly addresses the false negative problem in contrastive learning, where semantically equivalent items can appear as negatives and distort the learned geometry \cite{chuang2020debiased,robinson2021hard,kalantidis2020hard}. The batch objective is
\begin{align}
\mathcal{L}_{\text{MNR}} \;=\;
\frac{1}{|B|}\,
\sum_{i=1}^{|B|} \ell_i.
\end{align}

\textbf{Temperature and logit scale}
The temperature controls the classic alignment vs uniformity trade off in contrastive learning \cite{wang2020understanding}. We parameterize the MNR softmax with temperature $\tau$ and write the logit scale as $s{=}\tfrac{1}{\tau}$.
Larger $s$ sharpens the softmax and emphasizes harder negatives, typically enlarging margins but
increasing sensitivity to label or variant noise. Smaller $s$ softens negatives, improving
uniformity and robustness to heterogeneous query forms. Unless stated otherwise we use
$s{=}20$ with $\tau{=}0.05$.

\subsection{Query Free Real Time Operation}
Let $w_t$ denote the last $W$ seconds or $N$ turns of ambient conversation at time $t$. The query free variant encodes $w_t$ with the same tied Bi-encoder $f(\cdot)$ used elsewhere to obtain $c_t = f(w_t)$, then retrieves top $K$ orders by cosine similarity against precomputed order embeddings. No additional parameters, inputs, or objectives are introduced; this operational mode reuses the existing encoder and duplicate safe contrastive training. Short window conditioning also helps absorb conversational disfluencies and local ASR misrecognitions that would otherwise brittle a single utterance rewrite \cite{shriberg2001disfluencies,radford2023whisper}.

\section{Experimental Setup}

\subsection{Dataset and Temporal Split}
Experiments were conducted on a de identified corpus of clinical encounters containing provider patient conversations paired with their corresponding signed clinical orders. Each encounter was segmented into utterance level chunks and linked to canonical orderables spanning medications, laboratory tests, imaging, and procedures. GPT 5 was used as the LLM for constructing the training records, generating structured query to order alignments under variant constraints.

To emulate deployment conditions and prevent temporal leakage, encounters were divided chronologically into a six month training period and a one week test window. The training split comprised approximately 320{,}000 query to order pairs drawn from historical encounters preceding the cutoff date, while the test split contained roughly 1{,}400 unseen pairs from subsequent encounters. This temporal partitioning ensures that the model learns from past clinical behavior and is evaluated on genuinely future data.

\subsection{Model and Objective}
The base encoder is PubMedBERT, a biomedical transformer pretrained on PubMed abstracts and full text articles. It was fine tuned using the Multiple Negatives Ranking objective, which maximizes the similarity between each conversational query and its corresponding order while treating all other in batch positives as negatives. Duplicate safe batching prevents identical orders from appearing multiple times within a minibatch, ensuring valid negative sampling and avoiding trivial collisions.

Training was implemented in the SentenceTransformers framework with five epochs, a batch size of 64, and a learning rate of $2\times10^{-5}$ with a 0.1 warmup ratio. Inputs were truncated or padded to a maximum sequence length of 512 tokens, optimized using AdamW with linear decay. The scale value 20 was used for the MNR loss. All experiments ran in full precision on a single NVIDIA A100 GPU.

\subsection{Baselines and Open Embedding Models}
We compare the proposed model with both domain specific and general purpose embedding models to test whether reasoning regularized supervision provides benefits beyond model scale or instruction tuning. The primary baseline is PubMedBERT~\cite{neuml2020pubmedbert}, which serves as the domain encoder reference and initialization point for the proposed approach.

To contextualize against broader open source models, we evaluate several general purpose embedders distilled from large decoder backbones: Linq Embed Mistral~\cite{linqembedmistral2024}, SFR Embedding 2\_R~\cite{sfr2r2024}, SFR Embedding Mistral~\cite{sfrmistral2024}, GTE Qwen2 7B Instruct~\cite{gteqwen2_2024}, and GTE Qwen 1.5~\cite{gteqwen15_2024}. For encoder based comparators, we include BGE Large~\cite{bge2023} and Embedding Gemma~\cite{embeddinggemma300m2025}, representing strong, instruction tuned alternatives to domain pretraining.

\subsection{Evaluation Protocol}
Model performance was evaluated on the temporally held out test set using Recall@K and Mean Reciprocal Rank (MRR@K) for $K \in \{1,5,10,20\}$. Metrics were reported both in aggregate and across four linguistic variants, namely \emph{command only}, \emph{command+context}, \emph{context only}, and \emph{context+reasoning}, to measure robustness across different conversational expressions of clinical intent. In the query free framing, the retrieval input is a short rolling window of ambient dialogue; the same metrics and variant level stratification apply.

To approximate real deployment, we further evaluated an encounter scoped retrieval regime where each query is retrieved only against the candidate orders available within its corresponding encounter. This reflects realistic constraints such as service line, formulary, or clinician preferences. For each query, the retriever scored all encounter level candidates using cosine similarity. Two complementary evaluation modes were used, a \emph{strict} view, which assigns zero credit when the reference order is absent from the candidate pool, and a \emph{filtered} view, which restricts evaluation to queries whose reference order is present. The same metrics and variant level stratification were applied in both cases.

\subsection{Embedding Structure and Visualization}
To examine the learned embedding geometry, both quantitative and qualitative analyses were performed on a subset of 780 queries corresponding to 195 unique orders. Quantitative metrics included the mean margin and the fraction of positive margins, capturing discrimination between correct and hardest negatives, the mean compactness, which measures intra order similarity, the mean separation, which measures inter order centroid distance, the Fisher ratio, which measures the between to within cluster variance, and the cosine silhouette score \cite{rousseeuw1987silhouette}, which summarizes overall cluster cohesion. Together, these measures quantify how fine tuning reshapes the embedding space to achieve tighter alignment between conversational evidence and canonical orders.

Qualitative structure was visualized using UMAP \cite{mcinnes2018umap} with cosine distance, illustrating the spatial organization of query to query and query to order embeddings, per variant query to order similarity distributions, and histograms of inter order negative similarities. These visualizations complement the quantitative measures, revealing how the proposed model improves both local cohesion and global separability relative to the base encoder.

\subsection{Ablation Protocol}
To isolate the role of different supervision signals, four single variant models were trained under the same encoder, optimizer, and hyperparameter configuration as the main model, namely \emph{only command} ($q_{\text{c}}$), \emph{only context} ($q_{\text{ctx}}$), \emph{only command+context} ($q_{\text{cc}}$), and \emph{only context+reasoning} ($q_{\text{ctx+r}}$). Each model was trained exclusively on its respective variant, while evaluation used the mixed test set with the same per variant stratification as the main results. This setup enables both in variant performance comparison and assessment of cross variant generalization.

\textbf{Temperature sweep.}
We also ablate the MNR logit scale over $s\in\{10,20,100\}$ with all other hyperparameters unchanged, batch 64, learning rate $2{\times}10^{-5}$, warmup 0.1, max length 512, duplicate masking on.

\section{Results}

\subsection{Retrieval Performance}
We evaluate JEDA and PubMedBERT on a held out test set of 1{,}400 encounter derived queries using a unified orderable corpus. As shown in Table~\ref{tab:retrieval-main}, JEDA achieves substantial improvements across all ranks. Recall@1 rises from 0.08 to 0.28 (absolute +0.20), Recall@5 from 0.24 to 0.66 (absolute +0.42), and Recall@10/20 reach 0.82/0.91. Mean Reciprocal Rank improves by 2.5$\times$ (0.15 $\rightarrow$ 0.44), indicating stronger ranking calibration and steeper score separation near the top of the list.

Performance gains are consistent across query variants (Table~\ref{tab:retrieval-variant}). \emph{Command + context} shows the largest absolute improvement in R@1 (+0.252), followed by \emph{context + reasoning} (+0.203), \emph{command only (direct)} (+0.174), and \emph{context only (ambient)} (+0.157). The same ordering holds for MRR@10. This pattern suggests that JEDA leverages both explicit intent phrasing and supporting conversational evidence. When queries combine a directive with local context, retrieval is most reliable. When only ambient cues are present, performance still improves but remains the hardest setting.

\begin{table}[h!]
\centering
\caption{Overall retrieval performance.}
\label{tab:retrieval-main}
\begin{tabular}{lccccc}
\toprule
Model & R@1 & R@5 & R@10 & R@20 & MRR@10 \\
\midrule
PubMedBERT & 0.08 & 0.24 & 0.34 & 0.44 & 0.15 \\
JEDA       & 0.28 & 0.66 & 0.82 & 0.91 & 0.44 \\
\bottomrule
\end{tabular}
\end{table}

\begin{table}[t]
\centering
\caption{Variant level retrieval (R@1 and MRR@10).}
\label{tab:retrieval-variant}
\scriptsize
\setlength{\tabcolsep}{4pt}
\resizebox{\columnwidth}{!}{%
\begin{tabular}{lcc|cc}
\toprule
\multirow{2}{*}{Variant} & \multicolumn{2}{c}{PubMedBERT} & \multicolumn{2}{c}{JEDA} \\
 & R@1 & MRR@10 & R@1 & MRR@10 \\
\midrule
command + context                         & 0.084 & 0.158 & 0.336 & 0.501 \\
command only (direct)                     & 0.089 & 0.178 & 0.263 & 0.439 \\
context + reasoning (ambient + reasoning) & 0.079 & 0.144 & 0.282 & 0.440 \\
context only (ambient)                    & 0.068 & 0.121 & 0.225 & 0.368 \\
\bottomrule
\end{tabular}%
}
\end{table}

\subsection{Comparison with Open Embedding Baselines}
\label{sec:open-embedders}

To contextualize JEDA against state of the art open source embedding models, we evaluate several recent general purpose embedders distilled from large decoder backbones. Table~\ref{tab:retrieval-open} reports overall retrieval performance on the same held out test set used for PubMedBERT and JEDA.

\textbf{Overall performance.}
Among general purpose models, Linq Embed Mistral and SFR Embedding Mistral perform strongest, reaching R@1 of 0.12 to 0.13, followed by SFR Embedding 2\_R (0.07), BGE Large (0.09), and EmbeddingGemma 300M (0.05). Decoder derived models such as GTE Qwen2 7B Instruct and GTE Qwen 1.5 trail substantially despite their scale (R@1 $\leq$ 0.05 for the former and 0.048 for the latter). In contrast, JEDA (110 M parameters) achieves R@1 = 0.28, exceeding the next best by more than 2$\times$ while operating at encoder scale.

\begin{table}[h!]
\centering
\caption{Overall retrieval performance across open embedding baselines (regular test set).}
\label{tab:retrieval-open}
\footnotesize
\setlength{\tabcolsep}{4pt}
\begin{tabular}{lcccccc}
\toprule
Model & Params & R@1 & R@5 & R@10 & R@20 & MRR@10 \\
\midrule
PubMedBERT (base)         & 110M  & 0.08 & 0.24 & 0.34 & 0.44 & 0.15 \\
JEDA (ours)               & 110M  & 0.28 & 0.66 & 0.82 & 0.91 & 0.44 \\
\midrule
Linq Embed Mistral        & 7.1B  & 0.12 & 0.37 & 0.48 & 0.60 & 0.23 \\
SFR Embedding Mistral     & 7.1B  & 0.13 & 0.37 & 0.47 & 0.61 & 0.23 \\
SFR Embedding 2\_R        & 7.1B  & 0.07 & 0.26 & 0.40 & 0.52 & 0.15 \\
GTE Qwen2 7B Instruct     & 7B    & 0.05 & 0.10 & 0.12 & 0.15 & 0.07 \\
GTE Qwen 1.5              & 1.5B  & 0.048 & 0.091 & 0.107 & 0.134 & 0.067 \\
BGE Large (bge large en)  & 335M  & 0.087 & 0.272 & 0.381 & 0.461 & 0.171 \\
EmbeddingGemma 300M       & 300M  & 0.053 & 0.131 & 0.172 & 0.220 & 0.090 \\
\bottomrule
\end{tabular}
\end{table}

Table~\ref{tab:retrieval-open-variant} summarizes overall retrieval accuracy, while Table~\ref{tab:retrieval-open-variant} presents variant level R@1 performance across distinct query styles. JEDA consistently surpasses all open embedding baselines across every variant, with the largest margins for command+context and context+reasoning inputs, settings that most closely mirror real conversational orders. These results highlight that domain specific supervision not only improves overall recall but also yields more balanced robustness across directive, contextual, and reasoning rich formulations.

\begin{table}[h!]
\centering
\caption{Variant level retrieval across open embedding baselines (R@1).}
\label{tab:retrieval-open-variant}
\footnotesize
\setlength{\tabcolsep}{4pt}
\renewcommand{\arraystretch}{0.9}
\begin{tabular}{lcccc}
\toprule
Model & cmd+ctx & cmd-only & ctx+reason & ctx-only \\
\midrule
PubMedBERT (base)     & 0.084 & 0.089 & 0.079 & 0.068 \\
JEDA (ours)           & 0.336 & 0.263 & 0.282 & 0.225 \\
\midrule
Linq Embed Mistral    & 0.149 & 0.136 & 0.127 & 0.070 \\
SFR Embedding Mistral & 0.176 & 0.157 & 0.103 & 0.073 \\
SFR Embedding 2\_R    & 0.095 & 0.098 & 0.054 & 0.030 \\
GTE Qwen2 7B Instruct & 0.016 & 0.117 & 0.070 & 0.003 \\
GTE Qwen 1.5          & 0.030 & 0.081 & 0.062 & 0.019 \\
BGE Large             & 0.092 & 0.087 & 0.100 & 0.070 \\
EmbeddingGemma 300M   & 0.073 & 0.070 & 0.035 & 0.033 \\
\bottomrule
\end{tabular}
\end{table}

Decoder derived embedders, despite their scale and instruction tuning, underperform on conversational clinical intent retrieval. Their tokenization and optimization objectives, largely tuned for semantic similarity or generic question answering, do not capture fine grained reasoning cues that link patient context to order actions. JEDA uses duplicate safe contrastive supervision and reasoning enriched variants to produce sharper query to order alignment. These results indicate that domain aligned reasoning regularization can outperform billion parameter general models on downstream retrieval, even when trained at encoder scale.

\subsection{Embedding Structure and Geometric Properties}
\begin{figure}[h!]
\centering
\begin{minipage}{0.48\linewidth}
\centering
\includegraphics[width=\linewidth]{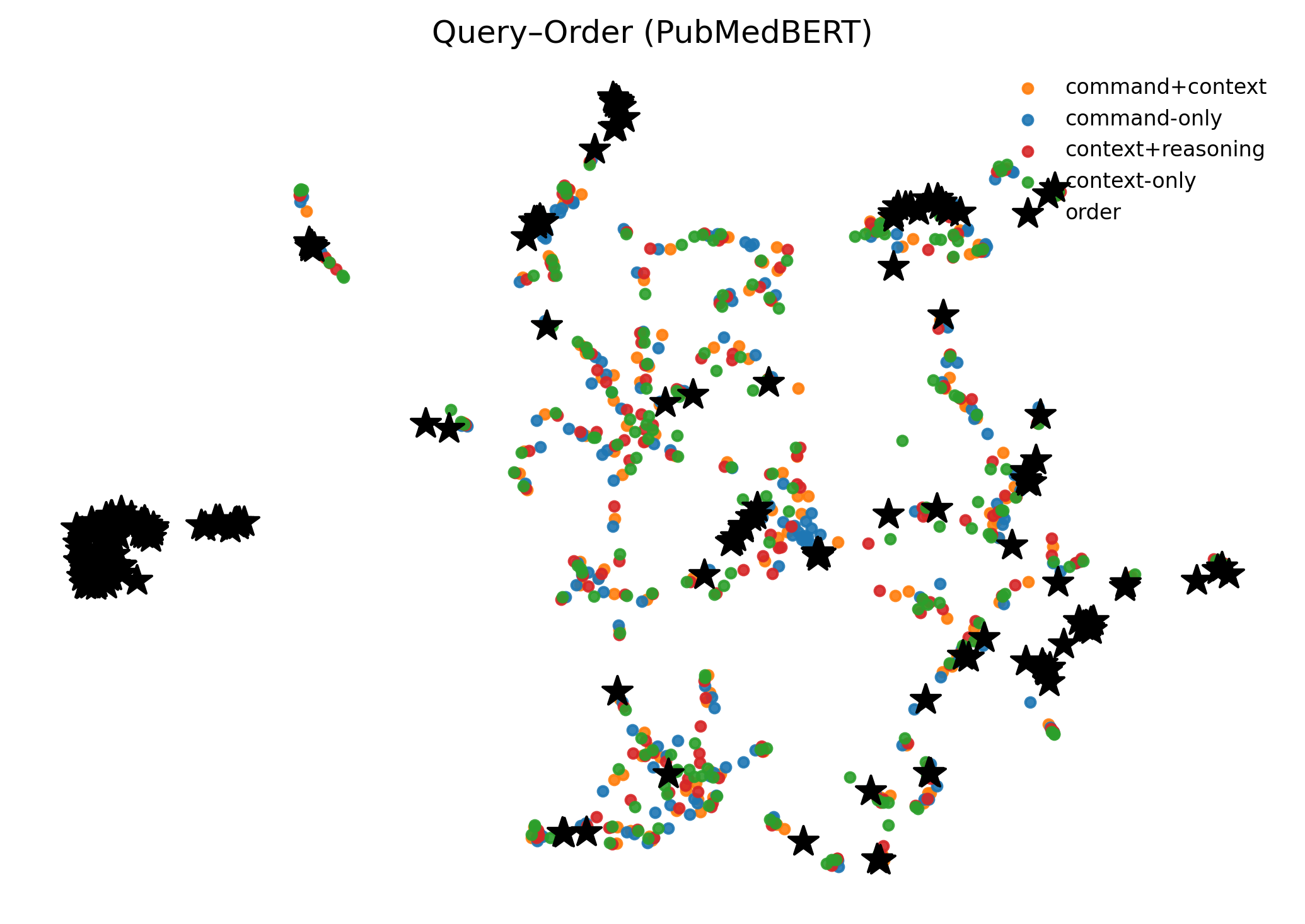}\\[-2pt]
\small (a) PubMedBERT
\end{minipage}\hfill
\begin{minipage}{0.48\linewidth}
\centering
\includegraphics[width=\linewidth]{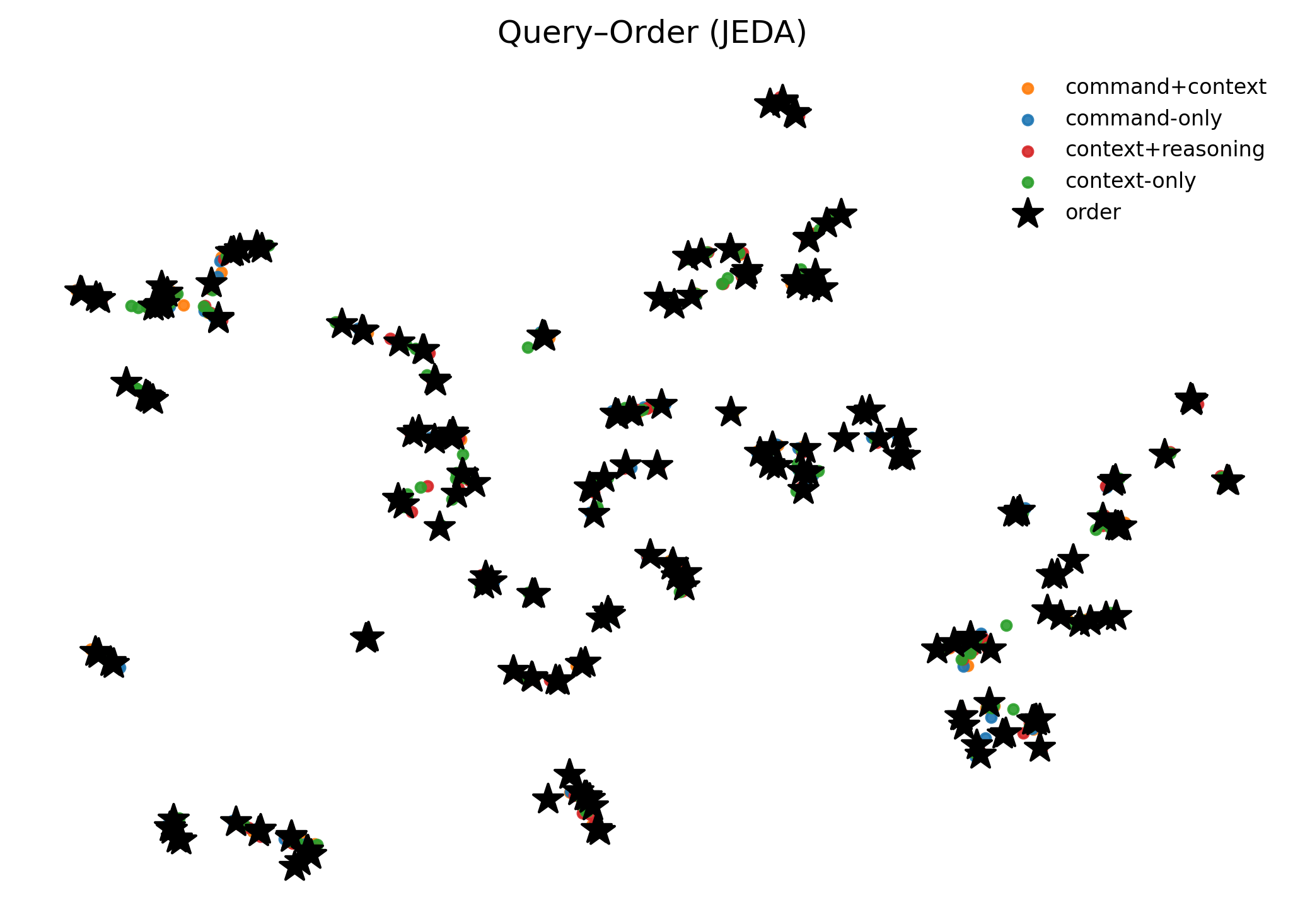}\\[-2pt]
\small (b) JEDA
\end{minipage}
\caption{Query to Order UMAP projections. JEDA produces tighter clusters around the correct order embeddings with clearer inter order separation compared to PubMedBERT.}
\label{fig:qo}
\end{figure}

Figure~\ref{fig:qo} visualizes the query to order structure for each model using UMAP with cosine distance. With PubMedBERT (Figure~\ref{fig:qo}a), queries are dispersed and form overlapping neighborhoods around order points, indicating weaker decision boundaries. JEDA (Figure~\ref{fig:qo}b) yields compact clusters that concentrate around the correct order embeddings with clearer inter order margins, reflecting a geometry that better supports top $k$ retrieval.

These qualitative observations align with the quantitative structure metrics in Table~\ref{tab:geometry-metrics} (780 queries, 195 unique orders). \emph{margin\_mean} increases from $-0.07$ to $0.12$ and the positive margin fraction from $0.30$ to $0.67$, showing that a majority of queries now score their true order above the strongest impostor. \emph{separation\_mean} rises from $0.20$ to $0.32$ and the \emph{Fisher ratio} more than doubles (0.96 $\rightarrow$ 2.24), indicating greater between order separation relative to within order spread. The \emph{silhouette} (cosine) improves from $0.28$ to $0.40$. Notably, \emph{compactness\_mean} remains essentially unchanged (0.087 $\rightarrow$ 0.088), suggesting that JEDA gains come primarily from enlarging inter order margins rather than over collapsing query clusters, which is consistent with better retrieval discrimination without sacrificing variant diversity.

\begin{table}[h!]
\centering
\caption{Embedding geometry metrics (780 queries; 195 unique orders).}
\label{tab:geometry-metrics}
\begin{tabular}{lcc}
\toprule
Metric & PubMedBERT & JEDA \\
\midrule
margin\_mean        & $-0.068$ & 0.125 \\
margin\_pos\_frac   & 0.299    & 0.673 \\
compactness\_mean   & 0.087    & 0.088 \\
separation\_mean    & 0.196    & 0.317 \\
fisher\_ratio       & 0.961    & 2.240 \\
silhouette\_cosine  & 0.279    & 0.399 \\
\bottomrule
\end{tabular}
\end{table}

\subsection{Variant Level Similarity and Negative Separation}
\begin{figure}[h!]
\centering
\begin{minipage}{0.48\linewidth}
\centering
\includegraphics[width=\linewidth]{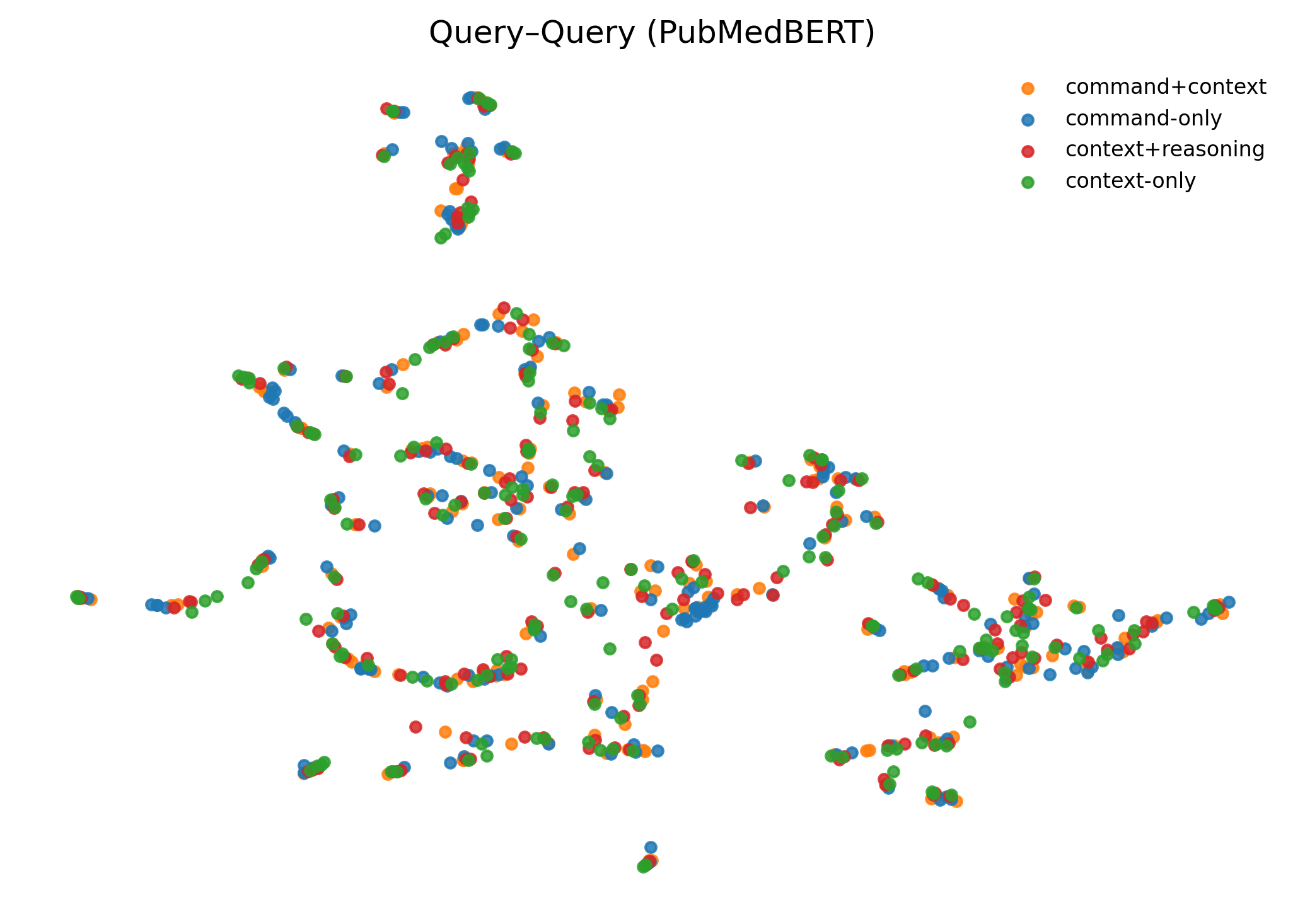}\\[-2pt]
\small (a) PubMedBERT
\end{minipage}\hfill
\begin{minipage}{0.48\linewidth}
\centering
\includegraphics[width=\linewidth]{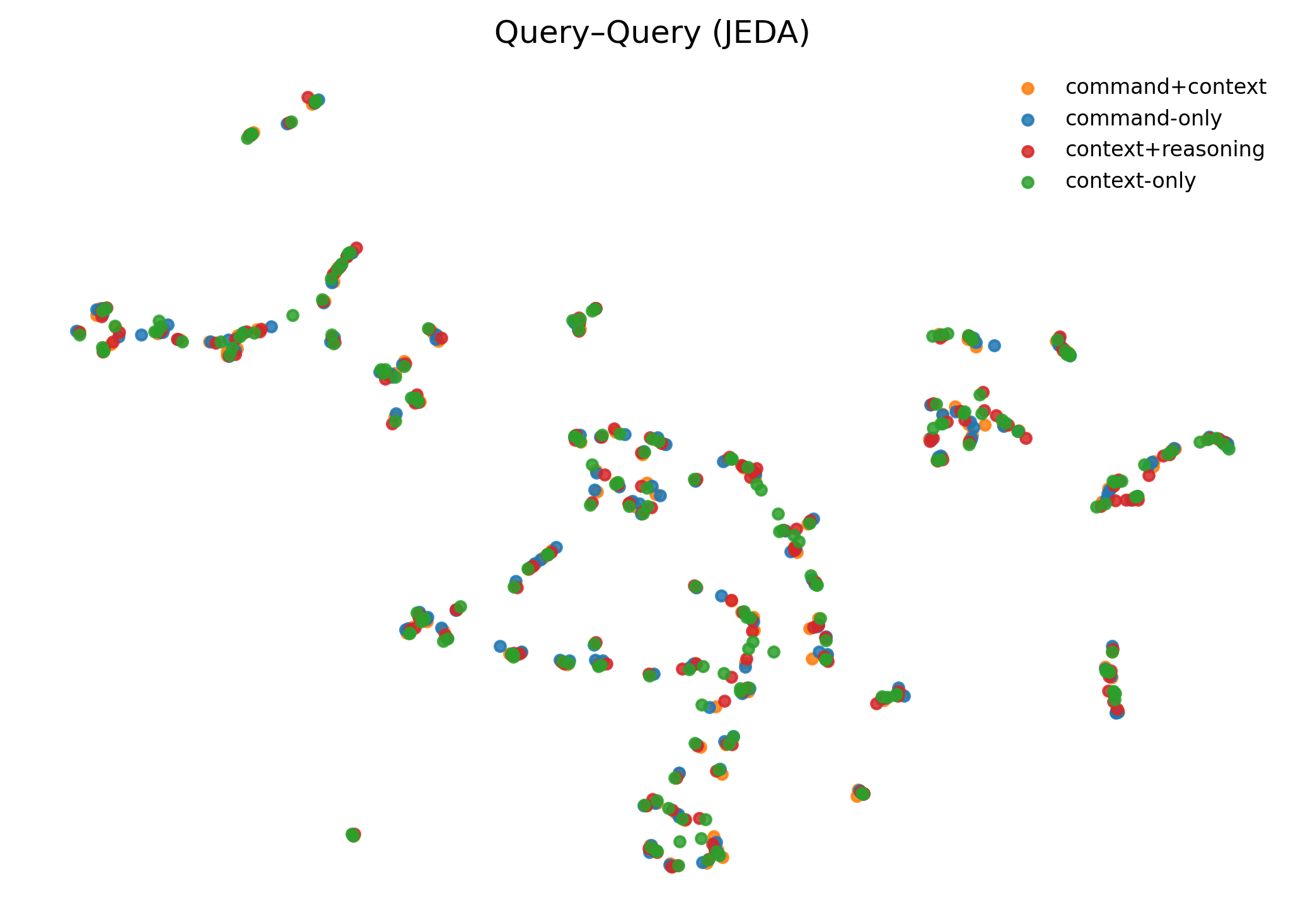}\\[-2pt]
\small (b) JEDA
\end{minipage}
\caption{Query to Query UMAP projections. JEDA exhibits tighter same order groupings and clearer boundaries.}
\label{fig:qq}
\end{figure}

\begin{figure}[h!]
\centering
\begin{minipage}{0.48\linewidth}
\centering
\includegraphics[width=\linewidth]{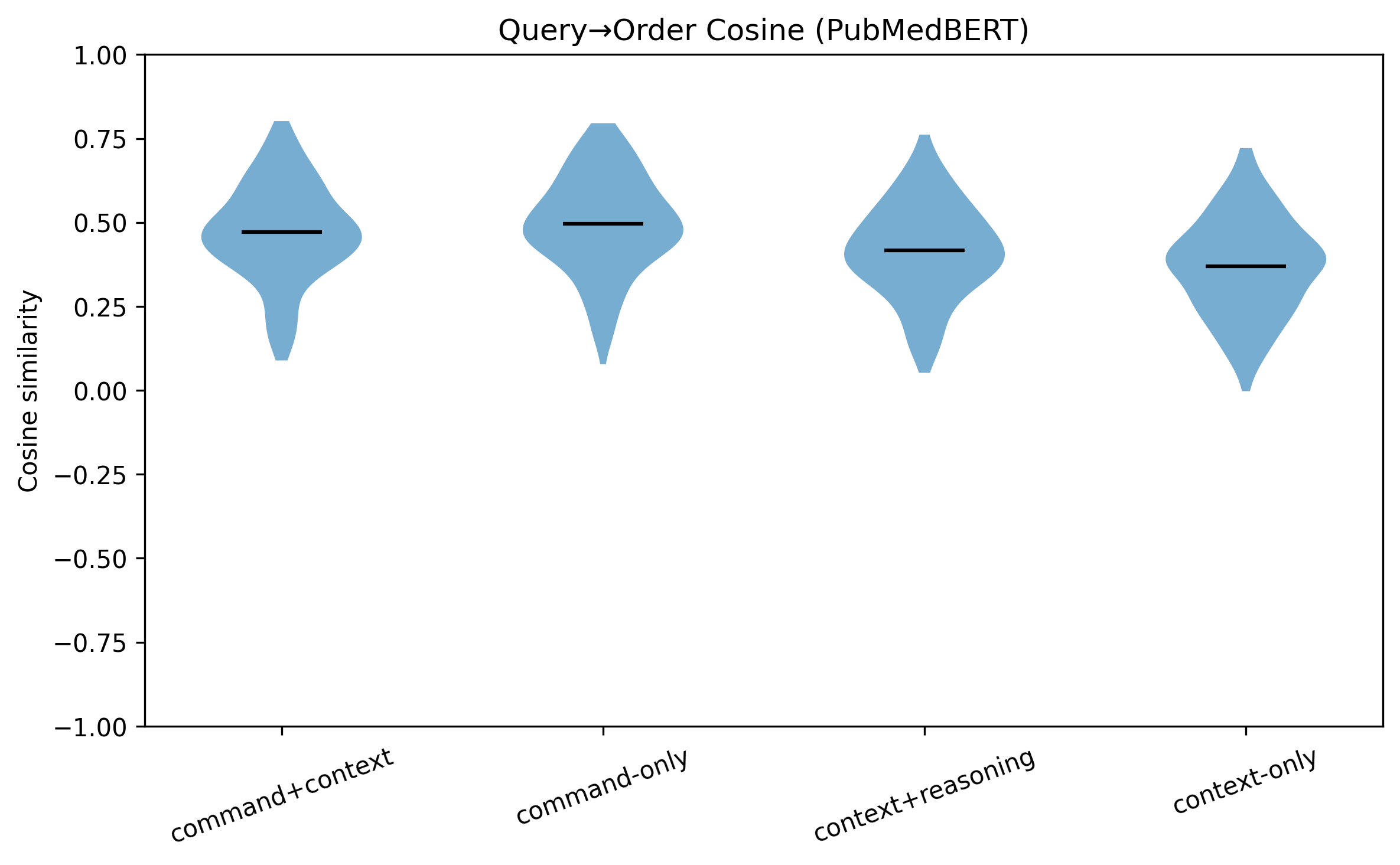}\\[-2pt]
\small (a) PubMedBERT
\end{minipage}\hfill
\begin{minipage}{0.48\linewidth}
\centering
\includegraphics[width=\linewidth]{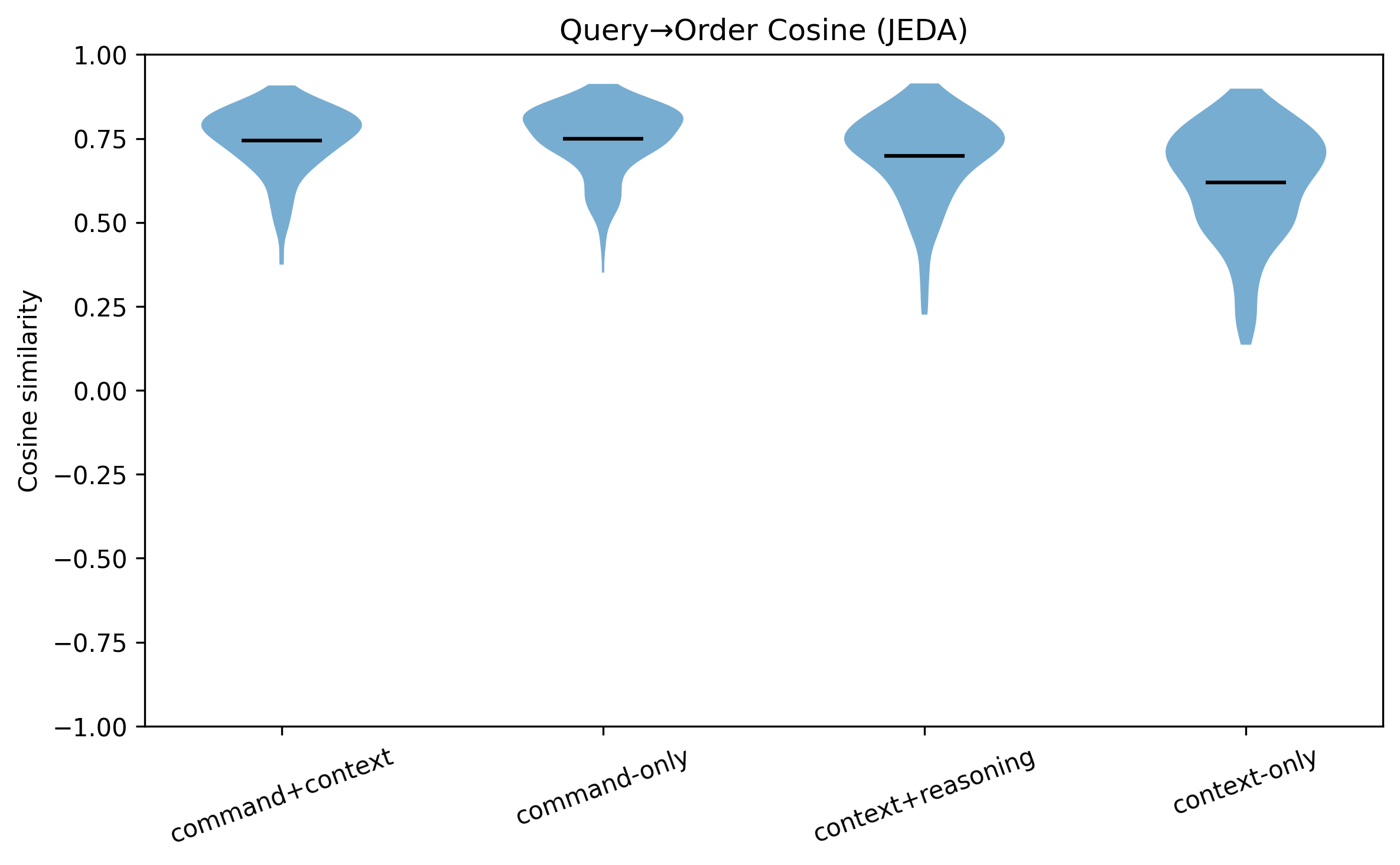}\\[-2pt]
\small (b) JEDA
\end{minipage}
\caption{Variant level Query to Order cosine similarity distributions. JEDA shows higher central tendency across variants.}
\label{fig:violin}
\end{figure}

\begin{figure}[h!]
\centering
\begin{minipage}{0.48\linewidth}
\centering
\includegraphics[width=\linewidth]{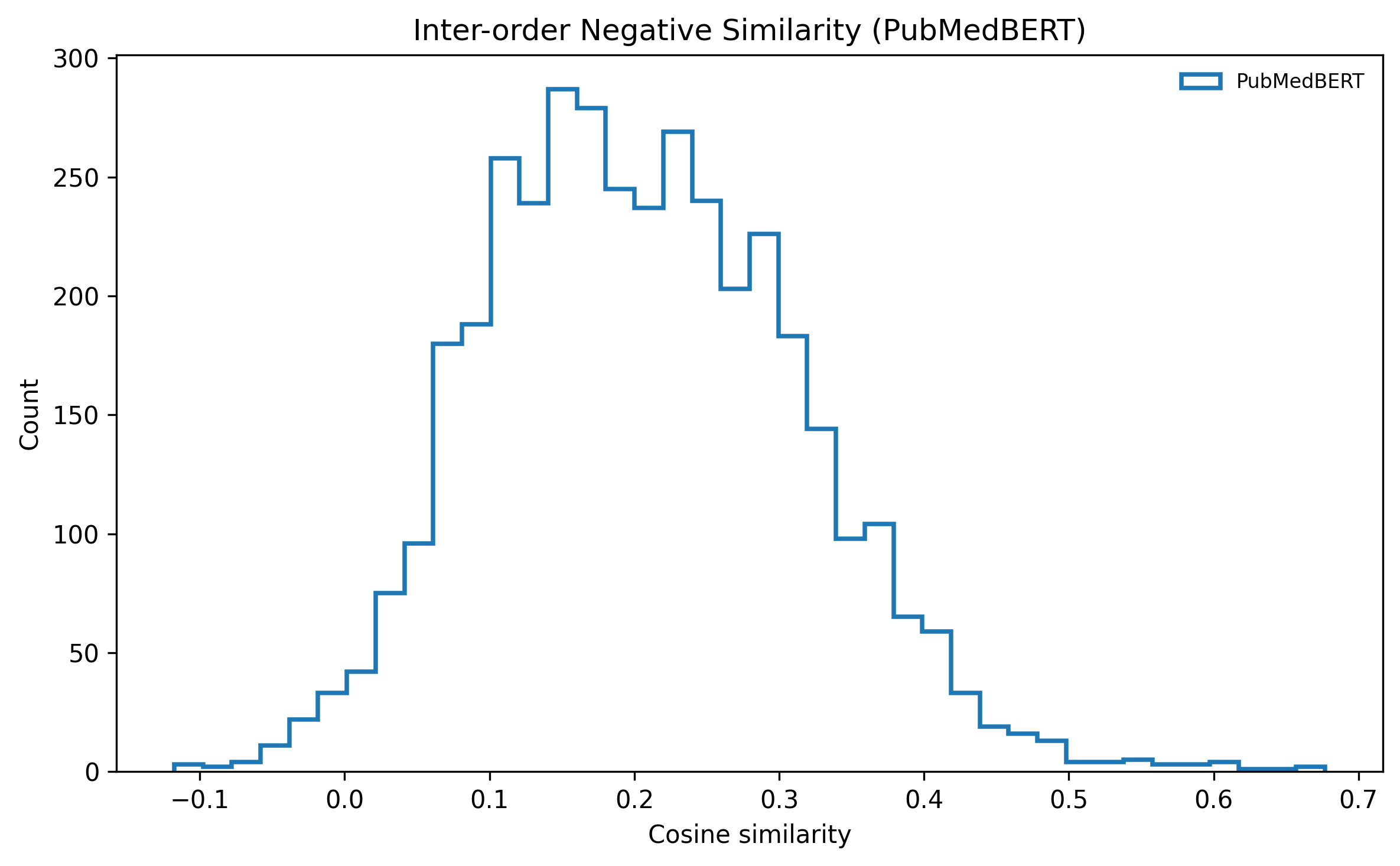}\\[-2pt]
\small (a) PubMedBERT
\end{minipage}\hfill
\begin{minipage}{0.48\linewidth}
\centering
\includegraphics[width=\linewidth]{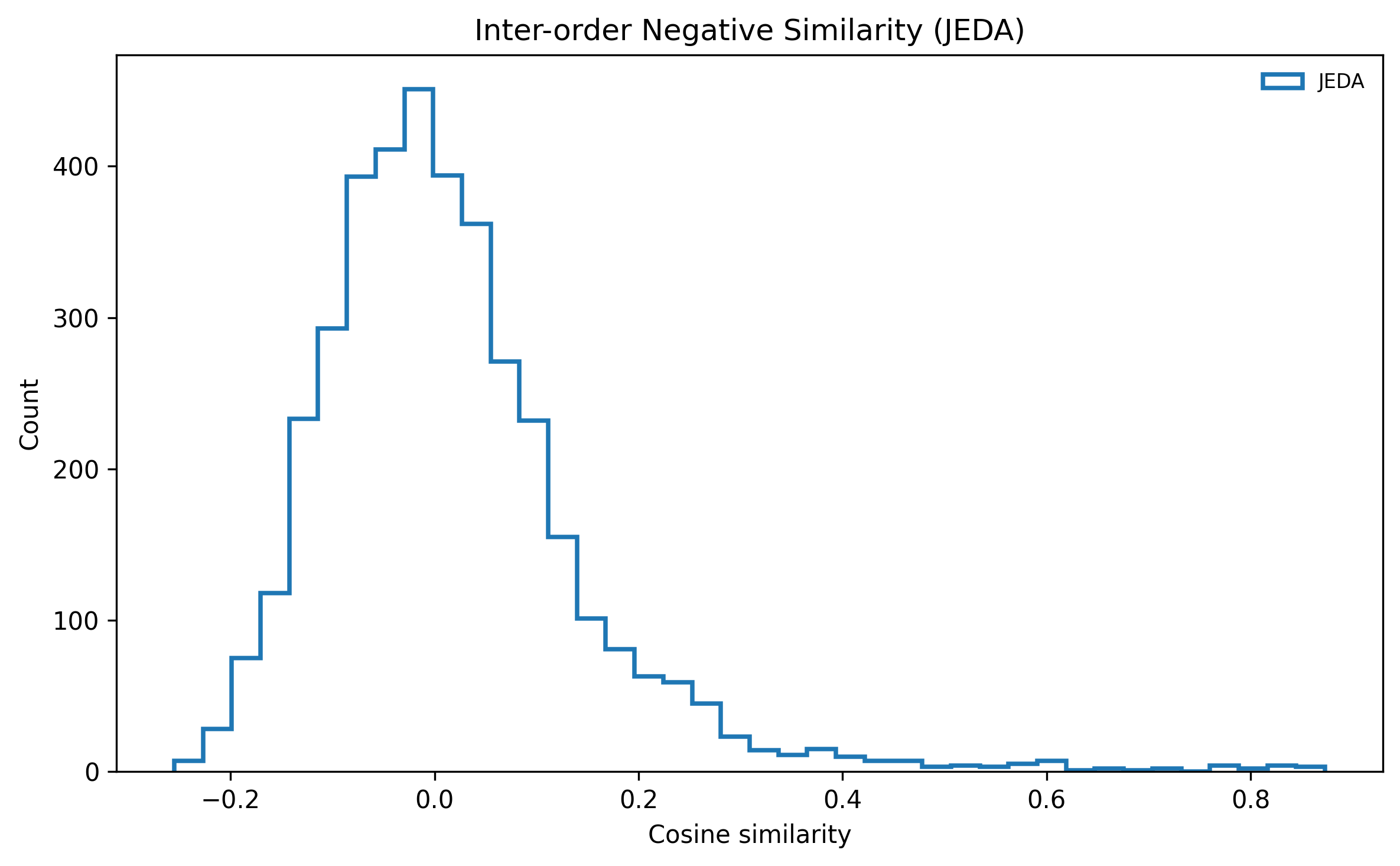}\\[-2pt]
\small (b) JEDA
\end{minipage}
\caption{Inter order negative similarity histograms. Under JEDA, unrelated pairs shift to lower cosine values, indicating improved discriminability.}
\label{fig:neg}
\end{figure}

\begin{figure}[h!]
\centering
\begin{minipage}{0.48\linewidth}
\centering
\includegraphics[width=\linewidth]{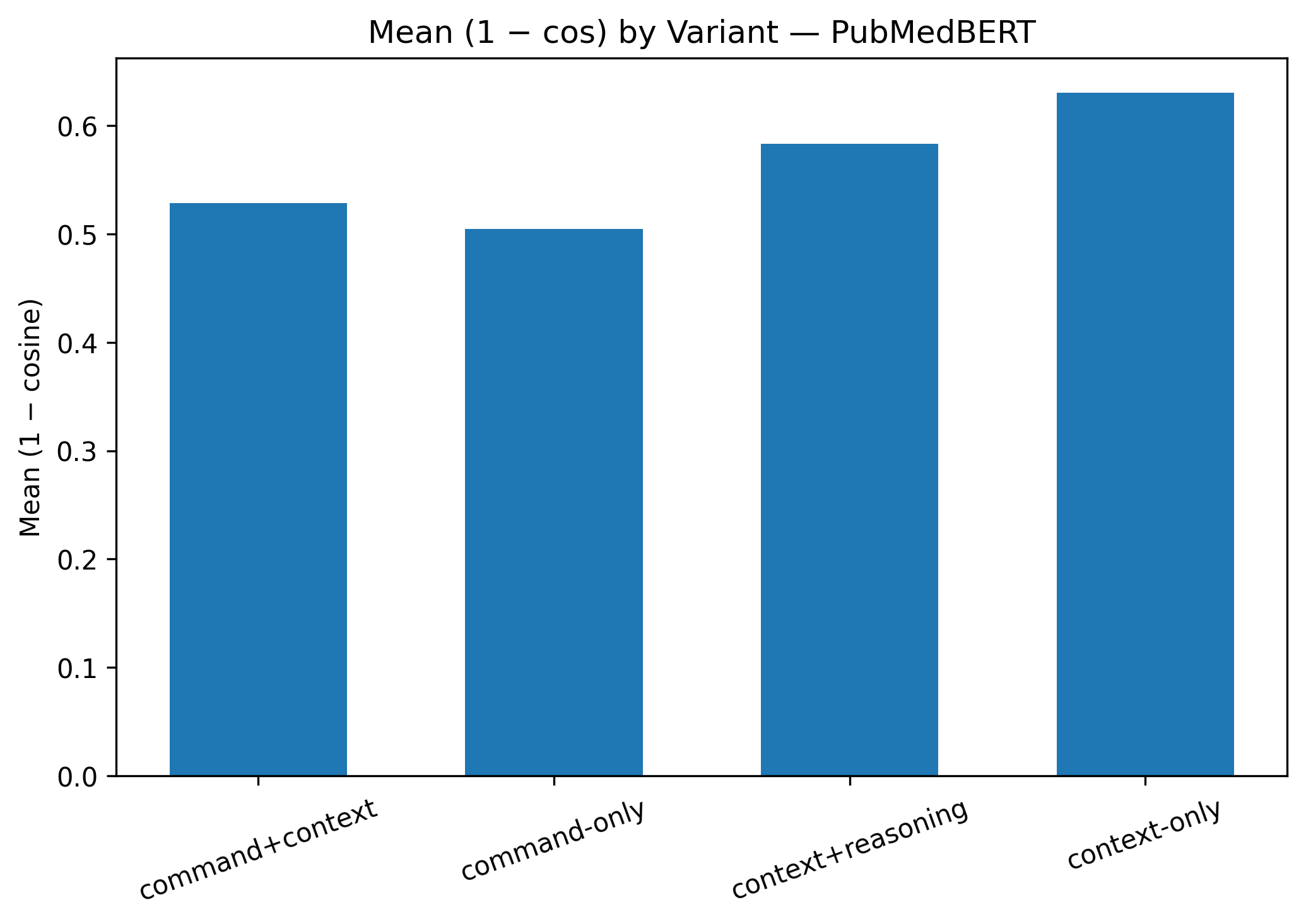}\\[-2pt]
\small (a) PubMedBERT
\end{minipage}\hfill
\begin{minipage}{0.48\linewidth}
\centering
\includegraphics[width=\linewidth]{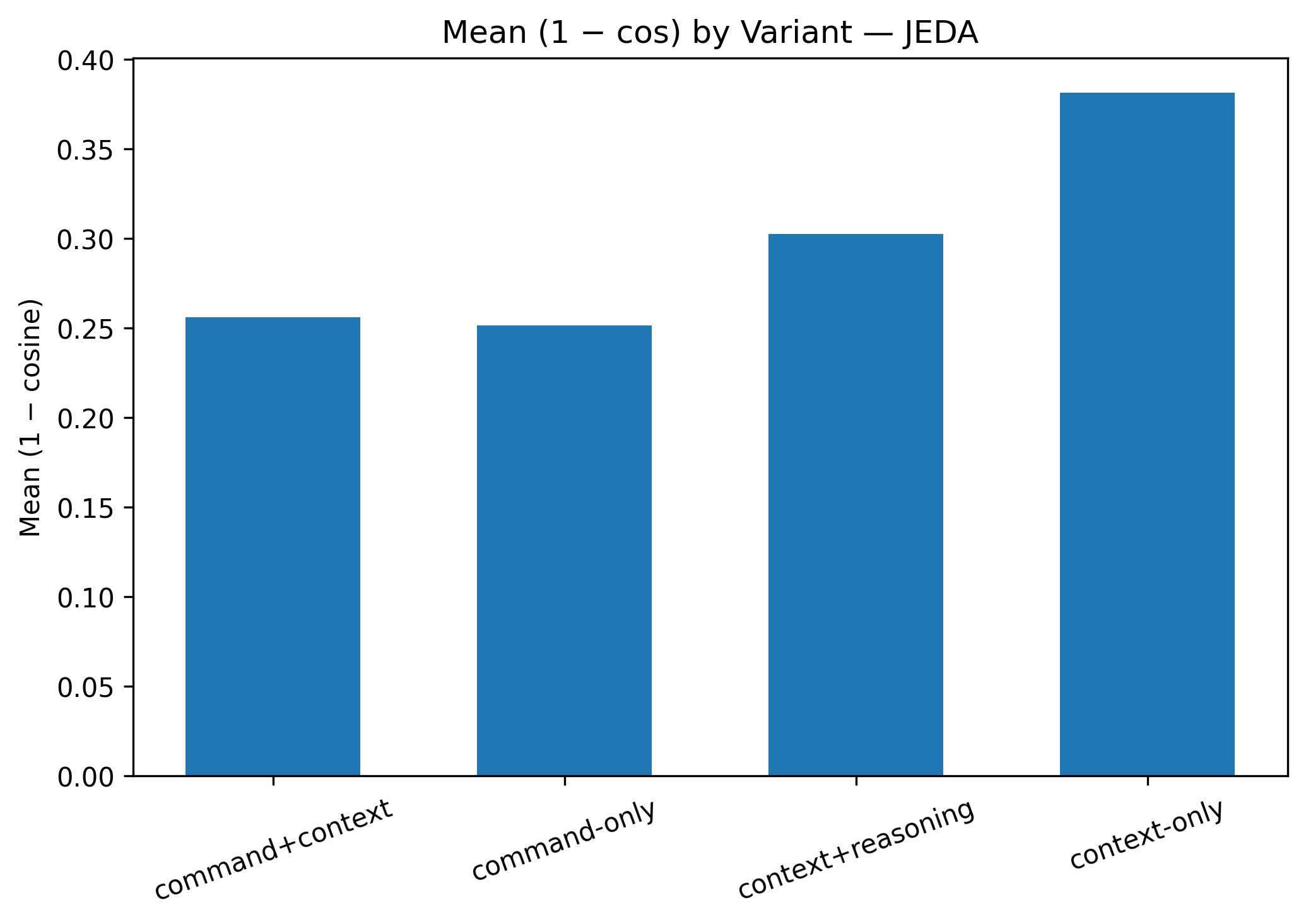}\\[-2pt]
\small (b) JEDA
\end{minipage}
\caption{Mean (1$-$cosine) by variant. Lower values under JEDA indicate tighter query to order coupling and reduced dispersion.}
\label{fig:one-minus-cos}
\end{figure}

Figures~\ref{fig:qq} to \ref{fig:one-minus-cos} provide complementary analyses. The query to query UMAPs (Figure~\ref{fig:qq}) show that JEDA reduces cross order mixing and yields crisper same order groupings. Variant level query to order cosine violins (Figure~\ref{fig:violin}) shift upward across all variants, with the largest lift for \emph{command + context} and \emph{context + reasoning}, matching the ranking improvements in Table~\ref{tab:retrieval-variant}. Inter order negative similarity histograms (Figure~\ref{fig:neg}) shift left under JEDA, indicating lower similarity to unrelated orders and stronger negative discrimination. Finally, mean (1$-$cosine) by variant (Figure~\ref{fig:one-minus-cos}) declines uniformly, confirming tighter query to order coupling while preserving variant structure.

Across ranking metrics, geometric diagnostics, and visual analyses, JEDA delivers consistent, sizable improvements over PubMedBERT. The model expands inter order margins without over collapsing intra order variation, resulting in a semantically coherent and discriminative embedding space that is resilient across direct command and ambient context formulations of clinical intent.

\subsection{Encounter Scoped Retrieval Under Context Constraints}

In many deployment scenarios, candidate orders are naturally constrained by encounter context such as service line, clinician preferences, and formulary policies. To approximate this setting, we evaluate retrieval with an encounter scoped candidate pool for each query. Two views are reported, a \emph{strict} view, which counts all test queries including those whose reference order is absent from the candidate pool, and a \emph{filtered} view, which conditions on queries where the reference is present. On the full set (N=1316), 160 queries lacked their reference in the pool, yielding a filtered set of N=1156. Results show that JEDA substantially outperforms PubMedBERT in both views, with larger gains in the filtered condition that isolates ranking quality independent of pool incompleteness.

Table~\ref{tab:ctx-overall-full} summarizes overall performance. In the strict view, JEDA improves Recall@1 from 0.084 to 0.433 and MRR@10 from 0.154 to 0.567. In the filtered view, JEDA reaches 0.493 R@1 and 0.645 MRR@10, reflecting strong top rank calibration when the true order is available. Tables~\ref{tab:ctx-variant-strict} and \ref{tab:ctx-variant-filtered} break down results by query variant. Improvements are consistent across \emph{command + context}, \emph{command only}, \emph{context + reasoning}, and \emph{context only}. The largest absolute gains appear for directive forms paired with local evidence, while purely ambient cues remain hardest yet still benefit. These trends align with our unconstrained corpus evaluation and the geometric analyses. JEDA tightens alignment of queries to their intended orders while increasing separation from confusable alternatives.

\begin{table}[t]
\centering
\caption{Encounter scoped retrieval (overall). \emph{Strict} counts all queries with missing references treated as errors, $N=1316$. \emph{Filtered} restricts to queries with the reference present, $N=1156$.}
\label{tab:ctx-overall-full}
\footnotesize
\setlength{\tabcolsep}{3pt}
\renewcommand{\arraystretch}{0.9}
\resizebox{\columnwidth}{!}{%
\begin{tabular}{lcccccccc}
\toprule
\multicolumn{9}{c}{\textbf{Strict ($N=1316$)}}\\
\midrule
Model & R@1 & R@5 & R@10 & R@20 & MRR@1 & MRR@5 & MRR@10 & MRR@20 \\
\midrule
PubMedBERT & 0.084 & 0.250 & 0.332 & 0.422 & 0.084 & 0.144 & 0.154 & 0.161 \\
JEDA       & \textbf{0.433} & \textbf{0.758} & \textbf{0.816} & \textbf{0.845} & \textbf{0.433} & \textbf{0.559} & \textbf{0.567} & \textbf{0.569} \\
\midrule
\multicolumn{9}{c}{\textbf{Filtered ($N=1156$)}}\\
\midrule
Model & R@1 & R@5 & R@10 & R@20 & MRR@1 & MRR@5 & MRR@10 & MRR@20 \\
\midrule
PubMedBERT & 0.096 & 0.285 & 0.378 & 0.480 & 0.096 & 0.164 & 0.176 & 0.183 \\
JEDA       & \textbf{0.493} & \textbf{0.863} & \textbf{0.929} & \textbf{0.962} & \textbf{0.493} & \textbf{0.636} & \textbf{0.645} & \textbf{0.648} \\
\bottomrule
\end{tabular}}
\end{table}

\textbf{Takeaways.}
(1) Encounter scoped pools raise the bar by forcing within context discrimination, and JEDA maintains strong margins in both strict and filtered views. (2) Variant level gains mirror the unconstrained setting. Explicit directives with local evidence benefit the most, yet ambient only queries still improve. (3) The gap between strict and filtered views, for example R@1 of 0.433 versus 0.493 for JEDA, quantifies the headroom available from better candidate generation and production filters. In practice, pairing JEDA with encounter aware rule based narrowing is expected to further boost top 1 precision.

\begin{table}[t]
\centering
\caption{Encounter scoped retrieval by variant (\emph{Strict}; $N_\text{strict}=329$ per variant).}
\label{tab:ctx-variant-strict}
\footnotesize
\setlength{\tabcolsep}{3pt}
\renewcommand{\arraystretch}{0.9}
\resizebox{\columnwidth}{!}{%
\begin{tabular}{lcccccccc}
\toprule
Variant & R@1 & R@5 & R@10 & R@20 & MRR@1 & MRR@5 & MRR@10 & MRR@20 \\
\midrule
\multicolumn{9}{l}{\emph{PubMedBERT}}\\
cmd+ctx & 0.100 & 0.258 & 0.362 & 0.468 & 0.100 & 0.160 & 0.173 & 0.181 \\
cmd-only & 0.103 & 0.368 & 0.471 & 0.578 & 0.103 & 0.194 & 0.207 & 0.215 \\
ctx+reason & 0.076 & 0.201 & 0.274 & 0.347 & 0.076 & 0.123 & 0.133 & 0.138 \\
ctx-only & 0.058 & 0.173 & 0.222 & 0.295 & 0.058 & 0.098 & 0.104 & 0.108 \\
\midrule
\multicolumn{9}{l}{\emph{JEDA}}\\
cmd+ctx & \textbf{0.517} & \textbf{0.818} & \textbf{0.848} & \textbf{0.863} & \textbf{0.517} & \textbf{0.639} & \textbf{0.643} & \textbf{0.645} \\
cmd-only & \textbf{0.447} & \textbf{0.809} & \textbf{0.848} & \textbf{0.860} & \textbf{0.447} & \textbf{0.586} & \textbf{0.591} & \textbf{0.592} \\
ctx+reason & \textbf{0.419} & \textbf{0.763} & \textbf{0.830} & \textbf{0.851} & \textbf{0.419} & \textbf{0.551} & \textbf{0.560} & \textbf{0.562} \\
ctx-only & \textbf{0.350} & \textbf{0.644} & \textbf{0.739} & \textbf{0.805} & \textbf{0.350} & \textbf{0.460} & \textbf{0.473} & \textbf{0.477} \\
\bottomrule
\end{tabular}}
\end{table}

\begin{table}[t]
\centering
\caption{Encounter scoped retrieval by variant (\emph{Filtered}; $N_\text{filtered}=289$ per variant).}
\label{tab:ctx-variant-filtered}
\footnotesize
\setlength{\tabcolsep}{3pt}
\renewcommand{\arraystretch}{0.9}
\resizebox{\columnwidth}{!}{%
\begin{tabular}{lcccccccc}
\toprule
Variant & R@1 & R@5 & R@10 & R@20 & MRR@1 & MRR@5 & MRR@10 & MRR@20 \\
\midrule
\multicolumn{9}{l}{\emph{PubMedBERT}}\\
cmd+ctx & 0.114 & 0.294 & 0.412 & 0.533 & 0.114 & 0.182 & 0.197 & 0.206 \\
cmd-only & 0.118 & 0.419 & 0.536 & 0.657 & 0.118 & 0.221 & 0.236 & 0.245 \\
ctx+reason & 0.087 & 0.228 & 0.311 & 0.394 & 0.087 & 0.140 & 0.151 & 0.157 \\
ctx-only & 0.066 & 0.197 & 0.253 & 0.336 & 0.066 & 0.111 & 0.118 & 0.123 \\
\midrule
\multicolumn{9}{l}{\emph{JEDA}}\\
cmd+ctx & \textbf{0.588} & \textbf{0.931} & \textbf{0.965} & \textbf{0.983} & \textbf{0.588} & \textbf{0.728} & \textbf{0.732} & \textbf{0.734} \\
cmd-only & \textbf{0.509} & \textbf{0.920} & \textbf{0.965} & \textbf{0.979} & \textbf{0.509} & \textbf{0.667} & \textbf{0.673} & \textbf{0.674} \\
ctx+reason & \textbf{0.478} & \textbf{0.869} & \textbf{0.945} & \textbf{0.969} & \textbf{0.478} & \textbf{0.627} & \textbf{0.638} & \textbf{0.640} \\
ctx-only & \textbf{0.398} & \textbf{0.734} & \textbf{0.841} & \textbf{0.917} & \textbf{0.398} & \textbf{0.524} & \textbf{0.538} & \textbf{0.543} \\
\bottomrule
\end{tabular}}
\end{table}

\subsection{Ablations}
\textbf{One Model per Query Variant}

Table~\ref{tab:abl-overall} summarizes overall retrieval for single variant training regimes, alongside PubMedBERT and the mixed JEDA model. The best single variant model is \emph{only context+reasoning} (R@1 0.226; MRR@10 0.377), narrowing the gap to mixed JEDA (R@1 0.276; MRR@10 0.437). Purely directive or purely ambient supervision performs worse, indicating that rationales provide information beyond raw context or keywords.

\begin{table}[h!]
\centering
\caption{Overall retrieval for single variant training (held out mixed test set).}
\label{tab:abl-overall}
\small
\begin{tabular}{lccccc}
\toprule
Model & R@1 & R@5 & R@10 & R@20 & MRR@10 \\
\midrule
PubMedBERT (base)      & 0.080 & 0.238 & 0.335 & 0.440 & 0.150 \\
JEDA (mixed variants)  & \textbf{0.276} & \textbf{0.661} & \textbf{0.816} & \textbf{0.912} & \textbf{0.437} \\
\midrule
only-command           & 0.196 & 0.568 & 0.707 & 0.799 & 0.351 \\
only-context           & 0.196 & 0.549 & 0.726 & 0.833 & 0.345 \\
only-command+context   & 0.205 & 0.556 & 0.724 & 0.806 & 0.354 \\
only-context+reasoning & \underline{0.226} & \underline{0.597} & \underline{0.764} & \underline{0.862} & \underline{0.377} \\
\bottomrule
\end{tabular}
\end{table}

Table~\ref{tab:abl-matrix} reports cross variant generalization (R@1). Two patterns emerge. Each single variant model peaks on its own evaluation style. The \emph{only context+reasoning} regime transfers best to other styles, especially ambient inputs, supporting the claim that explicit rationales regularize the space toward clinical logic rather than surface form.

\begin{table}[t]
\centering
\caption{Cross variant generalization (R@1). Rows are training regime. Columns are evaluation variant.}
\label{tab:abl-matrix}
\scriptsize
\setlength{\tabcolsep}{3pt}
\resizebox{\columnwidth}{!}{%
\begin{tabular}{lcccc}
\toprule
Training regime & cmd+ctx & cmd-only & ctx+reason & ctx-only \\
\midrule
only-command           & 0.252 & \textbf{0.279} & 0.133 & 0.122 \\
only-context           & 0.217 & 0.152 & \textbf{0.211} & 0.206 \\
only-command+context   & \textbf{0.320} & 0.225 & 0.157 & 0.119 \\
only-context+reasoning & 0.225 & 0.192 & \textbf{0.301} & 0.187 \\
\bottomrule
\end{tabular}%
}
\end{table}

\noindent\textbf{Interpretation.}
\emph{only-command} learns a strong mapping for directive phrasing but is brittle under ambient inputs. \emph{only-context} tolerates conversational noise but struggles on near neighbor confusions without explicit intent. \emph{only-command+context} benefits directives grounded in local evidence, yet still trails mixed training on ambient phrasing. \emph{only-context+reasoning} is the strongest single signal overall, consistently narrowing the gap to JEDA across variants. Reasoning turns supervision from “are related” to “are related because,” improving negative separation and robustness to underspecification. Mixing variants remains best. Exposure to intent, evidence, and rationales produces larger inter order margins and better top $k$ calibration than any single source.\\

\textbf{Temperature and Logit Scale Sensitivity}
We compare three logit scales for the MNR objective, $s{=}10$ with $\tau{=}0.1$, $s{=}20$ with $\tau{=}0.05$, and $s{=}100$ with $\tau{=}0.01$. Table~\ref{tab:temp-ablation} reports overall ranking. Table~\ref{tab:temp-geometry} reports embedding geometry. We observe that $s{=}10$ yields the strongest top $k$ metrics across mixed conversational variants, while $s{=}20$ attains the largest inter order separation. An extremely high scale, $s{=}100$, over hardens negatives, tightening intra order clusters but collapsing global separation and degrading ranking.

\begin{table}[h]
\centering
\caption{Temperature and logit scale ablation (overall ranking).}
\label{tab:temp-ablation}
\small
\begin{tabular}{lccccc}
\toprule
Scale $s{=}\tfrac{1}{\tau}$ & R@1 & R@5 & R@10 & R@20 & MRR@10 \\
\midrule
10  & \textbf{0.3089} & \textbf{0.7141} & \textbf{0.8333} & 0.9065 & \textbf{0.4744} \\
20  & 0.2800 & 0.6600 & 0.8200 & \textbf{0.9100} & 0.4400 \\
100 & 0.2392 & 0.6511 & 0.8232 & 0.9018 & 0.4123 \\
\bottomrule
\end{tabular}
\end{table}

\begin{table}[t]
\centering
\caption{Embedding geometry versus scale. Lower \textit{compactness} indicates tighter clusters.}
\label{tab:temp-geometry}
\setlength{\tabcolsep}{5pt}
\footnotesize
\begin{tabular}{lccc}
\toprule
\textbf{Metric} & \textbf{Scale 10} & \textbf{Scale 20} & \textbf{Scale 100} \\
\midrule
margin\_mean       & \textbf{0.163979} & 0.125000 & 0.029576 \\
margin\_pos\_frac  & 0.684615 & 0.673000 & \textbf{0.687179} \\
compactness        & 0.057952 & 0.088000 & \textbf{0.025411} \\
separation         & 0.280712 & \textbf{0.317000} & 0.085421 \\
Fisher             & \textbf{3.778496} & 2.240000 & 1.310661 \\
silhouette         & \textbf{0.436687} & 0.399000 & 0.426798 \\
\bottomrule
\end{tabular}
\end{table}

\noindent\textbf{Interpretation.}
Softening negatives with $s{=}10$ improves robustness across directive and ambient variants, raising R@1 and MRR and improving global geometry with higher Fisher and silhouette, while keeping clusters reasonably tight. The $s{=}20$ setting maximizes inter order centroid separation and remains strong on top 1 precision. Over hardening with $s{=}100$ collapses separation despite very tight clusters, which leads to worse ranking.


\section{Discussion}

\subsection{What the improvements reflect across baselines}
The retrieval gains in Table~\ref{tab:retrieval-main} align with the geometric changes in Table~\ref{tab:geometry-metrics}. In particular, the increase in \emph{margin\_pos\_frac} (0.30$\rightarrow$0.67), \emph{separation\_mean} (0.20$\rightarrow$0.32), and \emph{fisher\_ratio} (0.96$\rightarrow$2.24), alongside a higher \emph{silhouette} (0.28$\rightarrow$0.40), indicate that JEDA enlarges inter order margins without over collapsing variant diversity, with a nearly unchanged \emph{compactness\_mean}. The UMAP views in Figure~\ref{fig:qo} and Figure~\ref{fig:qq} visualize this effect. Order neighborhoods are more distinct and queries concentrate closer to their correct canonical order, supporting the observed gains in R@1 and MRR.

When extended to open embedding baselines (Table~\ref{tab:retrieval-open}), these geometric trends explain JEDA margin over decoder derived embedders such as Linq Embed Mistral and SFR Embedding Mistral. Despite their scale, those models exhibit diffuse neighborhoods and lower between order separation, confirming that domain aligned supervision and duplicate safe contrastive training are more effective than generic semantic pretraining for structured clinical retrieval.

\subsection{Encounter scoped retrieval and operational realism}
In an encounter scoped setting where each query is evaluated against only the orders relevant to that encounter, JEDA maintains strong advantages over PubMedBERT. Under the \emph{strict} view that counts all queries and treats missing positives as errors, overall R@1 improves from 0.084 to 0.433 and MRR@10 from 0.154 to 0.567. Under the \emph{filtered} view that considers only queries whose positives are in the candidate pool, overall R@1 rises from 0.096 to 0.493 and MRR@10 from 0.176 to 0.645. These results indicate that the geometric gains translate to a practical, constrained retrieval regime that mirrors production filtering.

\subsection{Direct and ambient expressions of intent}
Variant level outcomes show consistent improvements across \emph{command+context}, \emph{command only}, \emph{context+reasoning}, and \emph{context only}. In the encounter scoped \emph{filtered} setting, JEDA achieves R@1 of 0.588 (command+context), 0.509 (command only), 0.478 (context+reasoning), and 0.398 (context only), compared with PubMedBERT at 0.114, 0.118, 0.087, and 0.066, respectively. Notably, the lift for \emph{context only} (ambient) queries suggests that raw transcript spans are sufficient to trigger effective retrieval without an LLM query reformulator, simplifying the online path and reducing latency. The upward shifts in variant level query$\rightarrow$order similarity (Figure~\ref{fig:violin}) and the leftward shift of inter order negatives (Figure~\ref{fig:neg}) corroborate this behavior. This variant design is deliberate. Each form supplies a distinct supervisory bias, including intent isolation, evidence grounding, ambient robustness, and causal linkage, so alignment to the canonical order is preserved across the ways clinicians actually speak.

\subsection{Residual error modes}
Two patterns remain challenging. \textbf{(i) Fine grained, same family confusions:} High R@20 alongside lower R@1 indicates that the correct order often appears among top candidates but can be outranked by clinically proximate alternatives, for example closely related imaging protocols or medication formulation variants. Inter order negative histograms (Figure~\ref{fig:neg}) shift left under JEDA, yet a residual right tail persists for a subset of orders. \textbf{(ii) Sparse ambient cues:} \emph{context only} has the lowest absolute R@1 and MRR across variants. While it improves substantially with JEDA, broader cosine distributions remain when queries lack explicit directives.

\subsection{Production signals and re ranking}
Selecting the exact order typically benefits from encounter level signals such as visit type, department or service line, clinician favorites, patient constraints including allergies, active medications, and recent labs, and local formulary or ontology rules. In deployment, a lightweight, rule based filtering or deterministic re ranking pass over JEDA top $k$ can prioritize context consistent orders and suppress incompatible variants, which is expected to further improve top 1 recall while preserving Bi-encoder latency. Deployment should follow established best practices for clinical decision support to ensure usability and oversight \cite{bates2003tencommandments}.

\subsection{Practical implications}
The clearer separation observed in Figure~\ref{fig:qo} suggests more stable top $k$ neighborhoods under small input perturbations, which is important for synchronous, conversational workflows. Index size remains compact by storing a canonical representation per order. Organizations with heterogeneous nomenclature can optionally add a small set of curated synonyms to stabilize recall without materially affecting memory or throughput. The encounter scoped gains further indicate that JEDA benefits persist when retrieval is constrained to clinically plausible candidates.

\subsection{Deployment takeaways from ablations}
If constrained to a single supervision type, \emph{context+reasoning} yields the best ambient robustness and negative discrimination without inference time cost, since rationales are used only during training. For maximum reliability across mixed phrasing in real encounters, balanced mixing of \emph{command}, \emph{context}, \emph{command+context}, and \emph{context+reasoning} is preferred. The ablation matrix indicates this mixture best preserves cross variant invariance while enlarging inter order margins. Reasoning aware contrastive training underpins these results. By coupling ambient evidence with concise justifications during supervision and aligning all variants to the same canonical order, the objective learns invariances that reflect clinical logic rather than surface word overlap. In supervised contrastive terms, rationales provide structured signals that tighten within class alignment while maintaining uniformity across classes \cite{khosla2020supervised,wang2020understanding}. Duplicate safe masking further reduces false negative pressure from semantically equivalent items that would otherwise be treated as negatives \cite{chuang2020debiased,robinson2021hard,kalantidis2020hard}. The net effect is larger inter order margins, improved negative separation, and stronger robustness on ambient inputs, achieved without inference time cost because rationales are used only at training time. This mirrors classic findings that rationale supervision can steer models away from shortcut features toward task relevant evidence \cite{lei2016rationalizing,zaidan2007using}.

\subsection{Practical choice of temperature}
In our data, $s{=}10$ with $\tau{=}0.1$ offers the best end to end retrieval across mixed conversational inputs, while $s{=}20$ with $\tau{=}0.05$ provides wider class gaps that can benefit downstream encounter aware re ranking. We therefore recommend $s{=}10$ when recall and ambient robustness are primary, and $s{=}20$ when precise top 1 disambiguation is aided by production filters. Very high scales such as $s{=}100$ are not recommended.

\subsection{Contextualizing JEDA among open embedders}
JEDA performance relative to modern general purpose embedders demonstrates that domain specific reasoning supervision provides stronger inductive structure than scale alone. Models distilled from large decoders, such as Linq Embed Mistral or GTE Qwen2 7B Instruct, capture global semantic proximity but fail to distinguish clinically similar yet operationally distinct orders. By contrast, JEDA variant mixed training aligns queries and orders within a reasoning consistent manifold, producing higher recall and sharper negative separation at a fraction of the parameter count. This suggests that targeted supervision remains essential even as open embedding models continue to grow in size and generality.

\subsection{Limitations and next steps}
Our evaluation focuses on retrieval and embedding geometry, not downstream clinical impact such as time saved or error reduction. Residual errors concentrate among near neighbor order families and purely ambient queries. Future work includes ontology aware regularizers to further separate closely related orders, and modular re ranking for top $k$ verification, for example a small cross encoder or specificity rules, aiming to boost exact match precision while retaining JEDA latency profile. In parallel, benchmarking against emerging open domain embedders will continue to test whether architectural advances can close the gap without domain supervision.


\section{Conclusion}
\label{sec:conclusion}

Clinical ordering from natural conversation requires models that can bridge implicit reasoning with structured actions. JEDA addresses this gap by jointly embedding conversational fragments and canonical orders within a reasoning consistent space, eliminating the need for intermediate LLM query rewriting and enabling query free order search in real time. Through variant mixed contrastive training that balances directive, contextual, and causal supervision, JEDA achieves large gains in retrieval accuracy and geometric separation over its base model, PubMedBERT, as well as a diverse set of modern open embedders.

The improvements are not merely architectural but conceptual. Reasoning enriched supervision acts as a geometric regularizer that aligns clinical logic rather than surface semantics, and it increases noise resilience to underspecified or off target phrasing. The resulting model is lightweight, interpretable, and production ready, capable of real time retrieval with millisecond latency. When combined with encounter aware filtering, JEDA forms a transparent, efficient retrieval layer that directly connects ambient conversational evidence to actionable clinical orders.

Future directions include extending this formulation to multi institutional data, incorporating longer dialogue memory, and integrating JEDA embeddings into downstream reasoning and generation modules. More broadly, the results suggest that domain aligned reasoning remains a critical inductive bias, often surpassing sheer model scale, in bridging free form language with structured clinical decision support.


\section*{Ethics Statement}

\noindent\textbf{Data privacy and security.}
This study uses de identified, encounter derived text prepared under institutional privacy policies. No protected health information (PHI) was retained in the research artifacts, and no data left the secured computing environment. We did not attempt re identification. Access controls, encryption at rest, and audit logging were applied throughout. De-identification practices align with established methods and federal guidance \cite{neamatullah2008deid,dernoncourt2017deid,hhs2012deid}.

\noindent\textbf{Regulatory compliance.}
Data use complied with applicable regulations and institutional agreements. Where required, approvals were obtained under the institution’s governance processes. The work analyzes retrospective, de identified logs and does not intervene in patient care.

\noindent\textbf{Intended use and clinician oversight.}
JEDA is designed as a retrieval component to surface candidate orders; it is not intended to autonomously place orders or provide medical advice. In deployment, a licensed clinician remains in the loop and must confirm any recommendation before action. We explicitly discourage use outside supervised clinical workflows.

\noindent\textbf{Risk mitigation.}
To reduce the risk of harmful or inappropriate suggestions, we recommend pairing JEDA top $k$ outputs with rule based filters that enforce encounter context, patient specific constraints such as allergies, active medications, and recent labs, and local formulary policies, and with clear UI affordances for clinician review.

\noindent\textbf{Bias and fairness.}
Clinical text can reflect historical and institutional biases. We monitored performance across query variants, including direct versus ambient phrasing, and plan future audits on additional strata such as order types and care settings. We do not release patient data; any future model evaluations should include checks for disparate performance.

\noindent\textbf{Transparency and accountability.}
The system records provenance of retrieved candidates, for example order identifiers, to support auditing. We encourage sites to monitor post deployment behavior, including alert fatigue, acceptance rates, and near misses, and to establish feedback channels for corrective updates. Sites may also consider public facing documentation consistent with Model Cards for models and Datasheets for datasets \cite{mitchell2019modelcards,gebru2021datasheets}.

\noindent\textbf{Reproducibility and release.}
To protect patient privacy and institutional IP, we will \emph{not} release trained model weights or clinical datasets. We will provide code to reproduce all metrics, figures, and analyses on non sensitive corpora, along with detailed instructions to enable replication on appropriately governed institutional datasets.

\section{Acknowledgements}
We would like to thank other members of Oracle
Health AI for their collaboration while deploying JEDA, and Cody Maheu,  Shubham Shah, Praveen Polasam, Gyan Shankar, Ganesh Kumar,  Vishal Vishnoi, Raefer Gabriel and Kiran Rama for insightful feedback and discussions.

\bibliographystyle{ACM-Reference-Format}
\bibliography{software}

\end{document}